\pdfoutput=1

\documentclass[11pt]{article}

\usepackage[preprint]{acl}

\usepackage{soul}

\usepackage{color}
\usepackage{xcolor}
\definecolor{darkgreen}{RGB}{83,129,53}
\definecolor{darkred}{RGB}{163,21,21}

\usepackage{tikz}
\usepackage{subfig}
\usepackage{tcolorbox}

\usepackage{pgfplots}
\usepackage{wrapfig}
\usepackage{amssymb,mathrsfs}
\usepackage{amssymb}
\usepackage{array}
\usepackage{pgfplotstable}
\usepackage{pgf}
\usepackage[normalem]{ulem}
\usepackage{colortbl}
\usepackage{ulem}
\usepackage{xspace}

\usepackage{listings}

\usepackage{scalerel} 
\usepackage{times}
\usepackage{latexsym}
\usepackage{graphicx}
\usepackage{amsmath}
\usepackage{multirow}
\usepackage{amsthm}
\usepackage{amsfonts}
\usepackage{bm}
\usepackage{booktabs}
\usepackage{algorithm}
\usepackage{algorithmic}
\usepackage{verbatim}
\usepackage{tabularx}

\usepackage[T1]{fontenc}

\usepackage[utf8]{inputenc}

\usepackage{microtype}

\usepackage{inconsolata}

\newcommand{\KnowCoder}{\scalerel*{\includegraphics{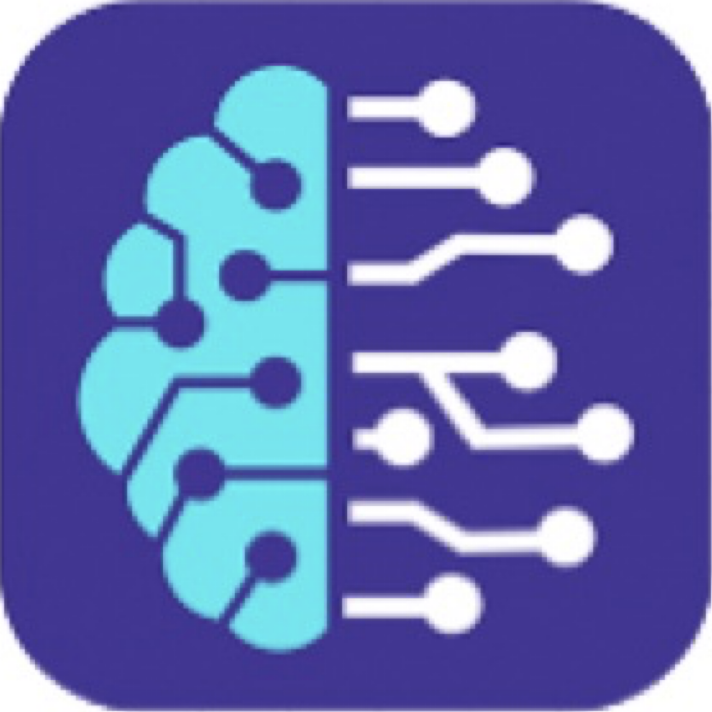}}{\textrm{\textbigcircle}}}

\usepackage{pifont}

\newcommand*\colourcheck[1]{%
  \expandafter\newcommand\csname #1check\endcsname{\textcolor{#1}{\ding{52}}}%
}
\colourcheck{blue}
\colourcheck{green}
\colourcheck{red}

\newcommand*\colourmark[1]{%
  \expandafter\newcommand\csname #1mark\endcsname{\textcolor{#1}{\ding{55}}}%
}
\colourmark{blue}
\colourmark{green}
\colourmark{red}

\definecolor{entity}{RGB}{113,172,213}
\definecolor{relation}{RGB}{132,174,110}
\definecolor{event}{RGB}{244,203,83}

\definecolor{dkgreen}{rgb}{0,0.6,0}
\definecolor{gray}{rgb}{0.5,0.5,0.5}
\definecolor{mauve}{rgb}{0.58,0,0.82}

\definecolor{softblue}{rgb}{0.88, 0.95, 1.0} 
\definecolor{softyellow}{rgb}{0.98, 0.98, 0.82} 

\sethlcolor{softblue}

\title{\raisebox{-.25\height}{\includegraphics[width=0.7cm]{pic/logo.png}} KnowCoder: Coding Structured Knowledge into LLMs for \\Universal Information Extraction}

\author{Zixuan Li\textsuperscript{}\thanks{\ \ Co-first authors.}\thanks{\ \ Corresponding
authors.}, Yutao Zeng\footnotemark[1], Yuxin
Zuo\textsuperscript{}\footnotemark[1], Weicheng
Ren\textsuperscript{}\footnotemark[1],\\  \textbf{Wenxuan Liu\textsuperscript{}, Miao
Su\textsuperscript{}, Yucan Guo\textsuperscript{}, Yantao
Liu\textsuperscript{}, Xiang Li\textsuperscript{}, Zhilei
Hu\textsuperscript{}, Long Bai\textsuperscript{},} \\
\textbf{Wei Li\textsuperscript{}, Yidan
Liu\textsuperscript{}, Pan Yang, Xiaolong
Jin\textsuperscript{}\footnotemark[2], Jiafeng Guo\textsuperscript{}\footnotemark[2], Xueqi Cheng\textsuperscript{}} \\
 \textsuperscript{}CAS Key Laboratory of Network Data Science and Technology, \\Institute of Computing Technology, Chinese Academy of Sciences\\
 \texttt{\{lizixuan, jinxiaolong, guojiafeng\}@ict.ac.cn}\\
 \url{https://ict-goknow.github.io/knowcoder/} \\[3pt]
 }

\begin{document}
\maketitle
\begin{abstract}

In this paper, we propose KnowCoder, a Large Language Model (LLM) to conduct Universal Information Extraction (UIE) via code generation. KnowCoder aims to develop a kind of unified schema representation that LLMs can easily understand and an effective learning framework that encourages LLMs to follow schemas and extract structured knowledge accurately. To achieve these, KnowCoder introduces a code-style schema representation method to uniformly transform different schemas into Python classes, with which complex schema information, such as constraints among tasks in UIE, can be captured in an LLM-friendly manner. We further construct a code-style schema library covering over $\textbf{30,000}$ types of knowledge, which is the largest one for UIE, to the best of our knowledge. To ease the learning process of LLMs, KnowCoder contains a two-phase learning framework that enhances its schema understanding ability via code pretraining and its schema following ability via instruction tuning. After code pretraining on around $1.5$B automatically constructed data, KnowCoder already attains remarkable generalization ability and achieves relative improvements by $\textbf{49.8\%}$ F1, compared to LLaMA2, under the few-shot setting. After instruction tuning, KnowCoder further exhibits strong generalization ability on unseen schemas and achieves up to $\textbf{12.5\%}$ and $\textbf{21.9\%}$, compared to sota baselines, under the zero-shot setting and the low resource setting, respectively. Additionally, based on our unified schema representations, various human-annotated datasets can simultaneously be utilized to refine KnowCoder, which achieves significant improvements up to $\textbf{7.5\%}$ under the supervised setting.

\end{abstract}

\section{Introduction}

\begin{figure}[tbp]  
    \vspace{8mm}
    \centering
    \includegraphics[width=1\linewidth]{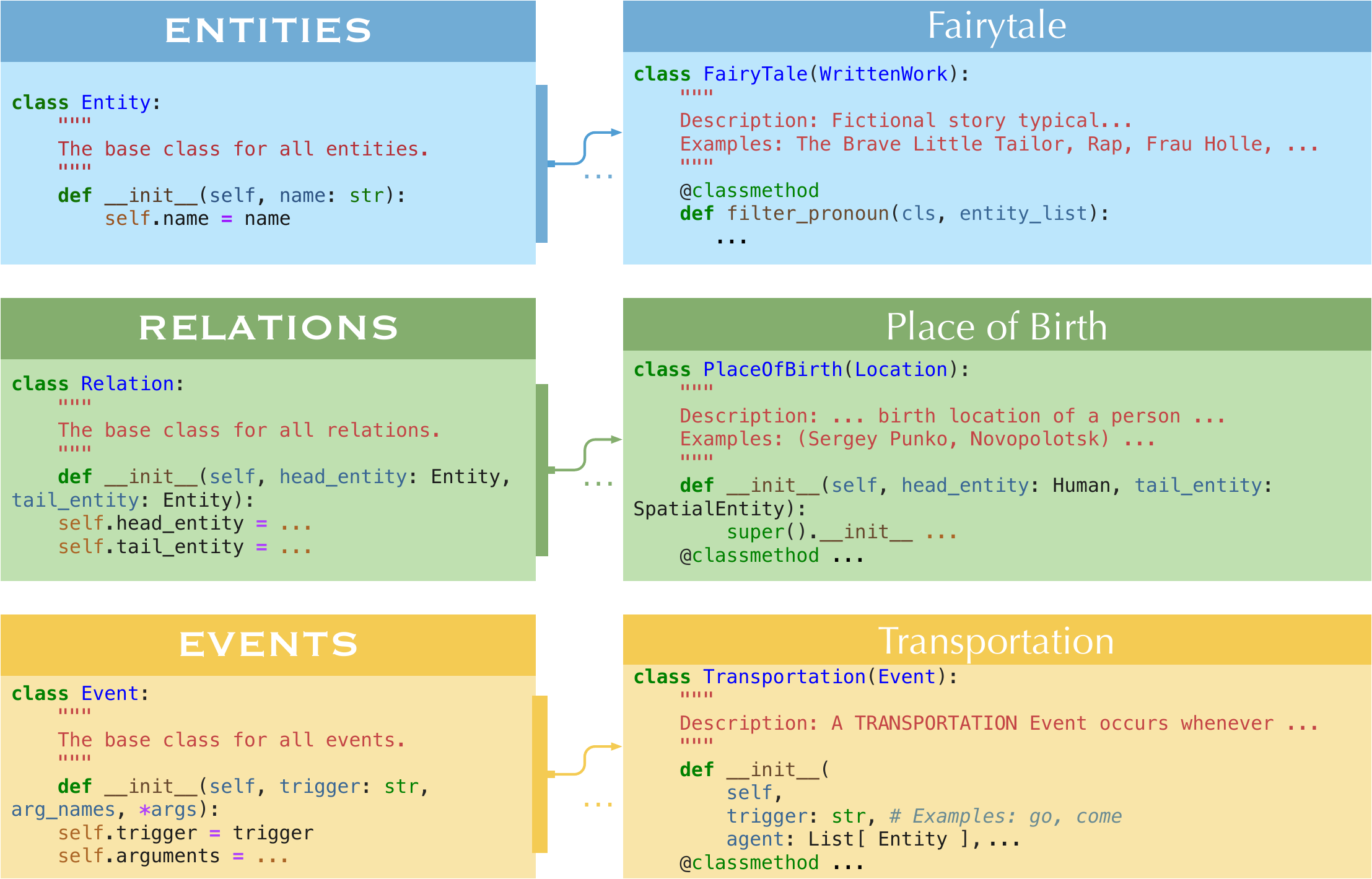}
    \caption{An illustration of KnowCoder schemas.} 
    \vspace{-4mm}
    \label{fig:illustration}
\end{figure}
  
Information Extraction (IE) aims to extract explicit and structured knowledge
following the manually designed schemas. The IE schemas define high-level types
of knowledge (i.e., concepts) and structures among them~\cite{kg-book}, which
include various types of entities, relations, and events. To simultaneously
extract various knowledge under different schemas via a single model, the
Universal Information Extraction (UIE) task is proposed~\cite{lin2020joint}.
Recently, Large Language Models (LLMs) have demonstrated general understanding
abilities through large-scale pretraining, which drives their increasing
utilization in UIE. However, their performance on UIE is still limited because
of two main challenges: (1) the lack of a unified schema representation method
that LLMs can easily understand; (2) the lack of an effective learning framework
that encourages LLMs to accurately follow specific schemas for extracting
structured knowledge.

For the first challenge, the existing UIE models first represent different
schemas in a universal way, such as classification labels~\cite{lin2020joint},
keywords~\cite{gui2023instructie}, or a specifically-designed formal
language~\cite{lu-etal-2022-unified}. These schema representation methods have
three main restrictions: (1) ignoring information like taxonomies (e.g.,
``fairytale'' is a subclass of ``written work'') and constraints among concepts
(e.g., ``spouse'' relation exists between two ``human'' entities); (2)
classification labels or a specifically designed formal language is hard for
LLMs to understand and follow; (3) designed for specific IE datasets and lacking
a general schema library.

To solve these restrictions, in this paper, we propose a kind of code-style
schema representation method, with which various types of knowledge are
generally defined as Python classes. As shown in Figure~\ref{fig:illustration},
the class inheritance mechanism is adopted to describe the concept taxonomies. A
mechanism of type hint is employed to model constraints among different
concepts. The class comments are used to provide clear definitions of concepts.
And, the class methods are used to post-process the results according to
specific IE guidelines. Upon this method, we construct a comprehensive
code-style schema library covering over $29,000$ entity types, $900$ relation types,
and $500$ event types based on Wikidata, the largest one for UIE, to the best of
our knowledge, currently reported in the open literature.

For the second challenge, the existing learning framework for UIE directly
conducts instruction tuning on LLMs to extract knowledge following specific and
limited schemas~\cite{sainz2023gollie,wang2023instructuie}. The enormous
concepts in the constructed schema library challenge the existing training
framework. To help LLMs better understand and follow these schemas, we propose
an effective two-phase framework containing a schema understanding phase and a
schema following phase. The former improves the ability of LLMs to understand
different concepts in schemas via large-scale code pretraining on the schema
definition code and corresponding instance code. The latter advances their
abilities to follow specific schemas in an IE task via instruction tuning. After code pretraining on around 1.5B automatically constructed data, KnowCoder already attains remarkable generalization ability and achieves NER improvements compared to the base model, LLaMA2, by $\textbf{49.8\%}$ relative F1 point under the few-shot setting on NER. After instruction tuning on 1.5B automatically annotated data,  KnowCoder experimentally demonstrates strong generalization ability on unseen schemas. Under the zero-shot setting,  KnowCoder achieves average relative improvements up to $\textbf{12.5\%}$ on  the NER task. Under the low-resource setting, KnowCoder gets average relative improvements up to $\textbf{21.9\%}$ on all the IE tasks. Additionally, based on our unified schema representation, various IE datasets can be simultaneously utilized to refine KnowCoder. After refinement, KnowCoder achieves consistent improvements across all IE tasks under the supervised setting, getting up to $\textbf{7.5\%}$ improvement on the relation extraction task, respectively. 

In general, the main contributions of this paper include: 
\begin{itemize}
  \item We propose a code-style schema representation method to uniformly
  represent different schemas for UIE. Using this method, we construct a large
  code-style schema library covering more than $30,000$ types of knowledge.
  \item We propose an effective learning framework for LLMs in a two-phase
  manner, which first enhances the schema understanding through code pretraining
  and then boosts schema following via instruction tuning. 
  \item After training on billions of automatically annotated data and refining
  with human-annotated IE datasets, KnowCoder demonstrates superior performance
  on different IE tasks under the zero-shot, low-resource, and supervised
  settings.
  \item The constructed schema library, training data, code, and models are released for future research.
\end{itemize}

\section{KnowCoder Schema} 
The proposed schema representation method uses code, a language that LLMs easy to understand, to define schemas. Specifically, KnowCoder schema adopts a series of programming language features to comprehensively model schema information, including the concept taxonomies, the constraints among different concepts, the definition of concepts, and other extraction requirements. Besides, considering that previous schema representation methods are only designed for specific datasets and contain limited types of knowledge, we further construct a
large-scale schema corpus containing a wide range of knowledge.

\subsection{Code-style Schema Representation Method}

 The code-style schema representation method comprises three basic classes,
 namely, \texttt{``\textcolor{darkgreen}{Entity}''}, \texttt{``\textcolor{darkgreen}{Relation}''}, and \texttt{``\textcolor{darkgreen}{Event}''}. Based on the three basic
 classes, we represent all the concepts in the schemas by the corresponding
 classes. Then, the instances of each concept can be represented by the objects
 of the corresponding class. In the following, we will introduce four features
 of the proposed representation method.

  \paragraph{Class Inheritance.} We adopt the class inheritance mechanism to
  account for the taxonomies in the schemas. Specifically, we let class A
  inherit all the class members from class B if the corresponding concept A is
  the hyponym of concept B in the taxonomies. For a concept with multiple
  hypernyms, the hypernym concept with the most instances is selected. The class
  of an unseen concept can inherit from an existing class or directly from the
  basic class.
  
  \paragraph{Class comment.} Similar to ~\citet{sainz2023gollie}, we adopt
  class comments to provide clear definitions of concepts. As shown in
  Figure~\ref{fig:illustration}, a class comment includes a natural language
  description that explains the corresponding concept and the examples of instances
  corresponding to that type. When there is an unseen concept, we use the
  description in its annotation guidelines~\footnote{If the annotation
  guidelines are missing, we use the description generated by GPT-4.} and
  manually give out a few examples. 

  \paragraph{Type Hint.} Type hint is a formal solution to indicate the type of
  a value in the code. We adopt type hints in the initialization function of a
  class to define its constraints with other classes strictly. Thus, the
  constraints among the concepts in the schemas are modeled. As shown in
  Figure~\ref{fig:illustration}, taking the relation ``PlaceOfBirth'' for
  example, \texttt{``def \_\_init\_\_(self, head\_entity: \textcolor{darkgreen}{Human}, tail\_entity:
  \textcolor{darkgreen}{SpatialEntity})''} denotes that the head entity must be a \texttt{``\textcolor{darkgreen}{Human}''} and the tail
  entity must be a \texttt{``\textcolor{darkgreen}{SpatialEntity}''}.

  \paragraph{Class Method.} A class method is bound to the class and not the
 object of the class. They are utilized to post-process the extracted instance
 results of a class. For example, some IE tasks may not consider the pronouns
 ``he'' and ``she'' as instances of the \texttt{``\textcolor{darkgreen}{Human}''} concept. To address this, a
 class method can be added to the \texttt{``\textcolor{darkgreen}{Human}''} class to filter out such pronouns
 from the extraction results, ensuring that the output aligns with the task's
 unique criteria. Note that, class methods are manually designed for specific IE
 tasks based on their task constraints. We take a few IE datasets to demonstrate
 the effectiveness of class methods in our experiments, as shown in the
 Appendix~\ref{sec:class_method}.

\subsection{Schema Library Construction}
We construct the code-style schema library based on Wikidata~\footnote{We use the Wikidata dump up to 20220704.}. We
select the concepts included in the existing IE datasets created from Wikidata, i.e., KELM~\cite{agarwal-etal-2021-knowledge}, UniversalNER~\cite{zhou2023universalner}, InstructIE~\cite{knowlm}, and LSEE~\cite{chen-etal-2017-automatically}. We derive the constraints among concepts according to their co-occurrences. To construct the taxonomies, we
extract the \texttt{``\textcolor{darkgreen}{SubclassOf}''} relations among these concepts from Wikidata. To obtain the description of a concept, we use its definition from Wikidata directly or generate its descriptions using GPT-4 if its definition in Wikidata
is missing. Finally, the constructed schema library encompasses over $29,177$ entity types, $876$ relation types, and $519$ event types. The detailed statistics of the schema are in Appendix \ref{appendix:stat}.

\begin{figure*}[tbp]  
  \centering
  \includegraphics[width=1.0\textwidth]{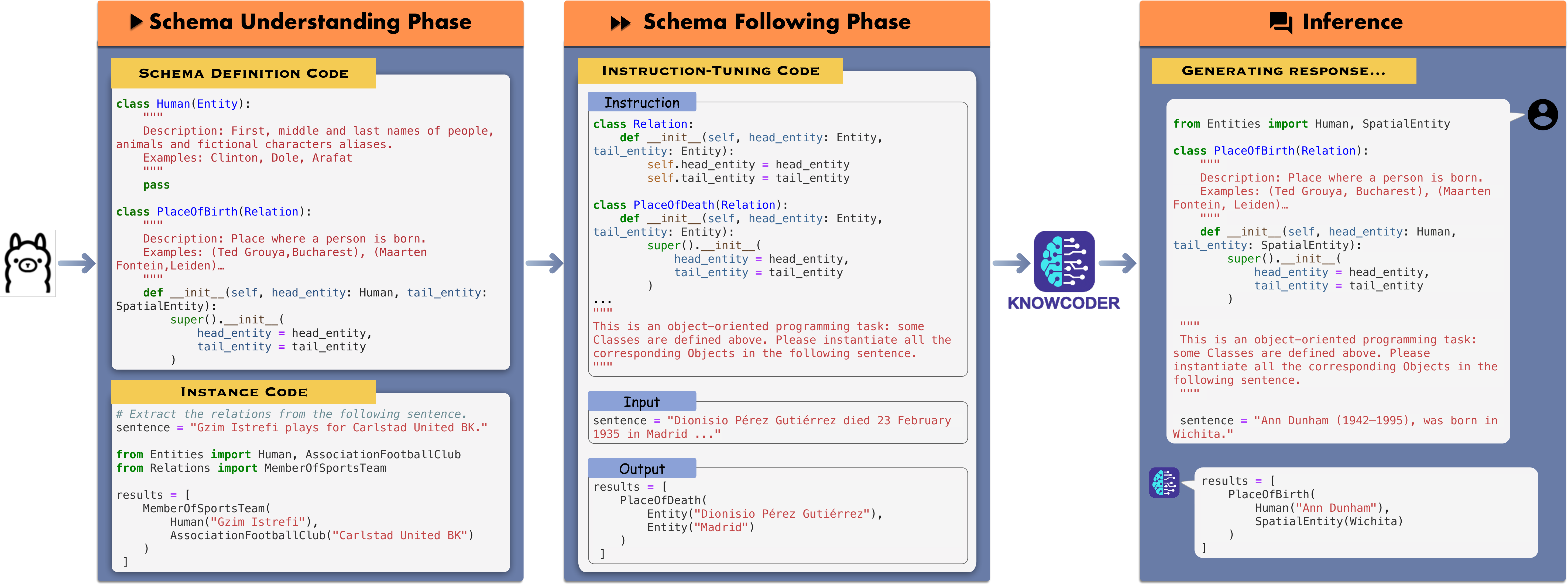}
  \caption{An diagram of training and inference processes of KnowCoder.}
 \vspace{-4mm}
  \label{fig:training_framework}
  \end{figure*}

\section{Learning Framework of KnowCoder}

To discriminate enormous concepts defined in schemas, we first let KnowCoder understand each concept through its definition and instances. Subsequently, we enhance KnowCoder to discriminate among a few concepts and extract corresponding knowledge. Thus, as shown in Figure~\ref{fig:training_framework}, the proposed learning framework contains two phases, i.e., the schema understanding phase and the schema following phase. In the schema understanding phase, KnowCoder undergoes code pretraining to understand each concept in two manners: 1) Go through the class definition code of each concept. 2) Go through the instance codes of each concept. In the schema following phase, KnowCoder is finetuned using instruction tuning code, where multiple task-demanded concepts are given in the schemas, enhancing KnowCoder's ability to follow schemas and generate instantiating code accordingly.

\subsection{Schema Understanding Phase}

\subsubsection{Training Data Generation}
To enhance KnowCoder's schema understanding abilities, we construct a large-scale training dataset based on the schema library. As shown in the left part of Figure~\ref{fig:training_framework}, the training data consists of two kinds of codes, i.e., schema definition codes and instance codes. The schema definition codes are generated based on the schema library, where we randomly sample a certain number of concepts (decided by the maximum sequence length) from the schema library to consist of a training sample. As the aim of the schema understanding phase is to understand each concept but not to discriminate various concepts, the instance code corresponding to a single concept contains three parts, i.e., a sentence containing instances of the given concept, an import clause to introduce the corresponding class of the given concept, and an instantiating clause to give out all the instances of the given concept in the sentence. The schema-instance codes are constructed based on KELM corpus~\cite{agarwal-etal-2021-knowledge}, which contains $15, 628, 486$ synthetic sentences to describe the structured knowledge from Wikidata. We do data cleaning for the corpus. The cleaning details are in Appendix \ref{sec: data_clean}.

\subsubsection{Code Pretraining}

After obtaining the data, we apply regular code pretraining to make LLM understand the diverse concepts in the schemas. Given a training sample with length of $L$, $X = {x_0, x_1, ..., x_{i}, ..., X_{L-1}}$, the model attempts to predict every token $x_{l}$ based on the ${x_0, ..., x_{l-1}}$, where $l={0,...,L-1}$. Some training details are as follows:

\paragraph{Schema Importing.}
The straightforward way to construct a pretraining sample is to directly give the whole schema definition for the corresponding instance code. However, this manner may cause the model to overfit the schema definition code because they are frequently repeated in every instance code. To address this problem, we separate the schema definition code from the instance code and use the ``import'' clause to introduce the corresponding schema definition to the instance code.

The position of the ``import'' clause is also critical for the LLMs to learn. We study two positions for the ``import'' clause, i.e., ``Import-First'' and ``Sentence-First''. We adopt ``Sentence-First'' in the learning framework because it performs better than the others. The comparison results are in Appendix~\ref{sec:import_position}.

\subsection{Schema Following Phase}

\subsubsection{Training Data Generation}

  To enhance the schema following abilities of KnowCoder, we construct instruction tuning training data for UIE tasks. As shown in the middle part of Figure~\ref{fig:training_framework}, a typical instruction tuning sample contains three parts of codes, i.e., instruction code $T$, input code $I$, and output code $O$. 
  
  The instruction code $T$ comprises two snippets, i.e., schema definition and task description. The schema definition snippet includes definitions of some concepts selected from the former phase, which defines specific concepts to be extracted. The task description snippet includes a comment that contains a natural language description of an IE task. For example, the task description of Relation Extraction (RE) is ``This is an object-oriented programming task: some Classes are defined above. Please instantiate all the corresponding Objects in the following sentence.''. The input $I$ contains the sentence to be extracted, which is denoted as a variable ``sentence'', i.e., ``sentence = ...''. The output $O$ contains all the golden knowledge in the sentence, denoted as a list variable ``results'', i.e., ``results = [...]''. We have conducted a performance comparison of different versions of the instructions, and the corresponding results are in Appendix~\ref{sec:prompt}.
  
  We construct the training corpus from three data sources. For Named Entity Extraction (NER), ChatGPT-annotated Pile corpus~\cite{zhou2023universalner} is selected. For Relation Extraction (RE) and Event Extraction (EE), we adopt the data sources constructed in \citet{gui2023instructie}\footnote{We use the English version of the constructed data source.} and LSEE~\cite{chen-etal-2017-automatically}, respectively.

\subsubsection{Instruction Tuning}

The objective of instruction tuning is to learn an LLM $\mathbf{f}:(I \times T) \rightarrow O$. The LLM takes input code $I$, and instruction code $T$ as input. Subsequently, the LLM is tuned to generate every token in the output $O$. Some training details are as follows:

\paragraph{Negative Class Sampling.} In the constructed schema library, there are more than $30000$ concepts. It is challenging for the model to accommodate all the corresponding class definitions in a single prompt. Consequently, KnowCoder employs a negative class sampling strategy. For each training sample, in addition to the classes annotated in the sentence, we randomly sample several classes ($20\%$ number of the golden classes) from the remaining classes.

\paragraph{Fully negative Sample Construction.} In real-world scenarios, many sentences do not contain any knowledge of a specific IE task, called fully negative samples in this paper. However, the selected data sources neglect such samples. To address this problem, we randomly sample $5\%$ sentences from the data sources. For each sentence, we replace the golden classes with five random negative classes.

\subsection{Refinement}

After schema understanding and following, we obtain KnowCoder, an LLM that demonstrates strong generalization ability on unseen schemas. Additionally, based on our unified schema representation, KnowCoder can be further refined by various human-annotated datasets simultaneously. In this phase, we conduct instruction tuning based on the datasets used in previous work~\cite{wang2023instructuie, sainz2023gollie}.


In different IE datasets, concepts with the same name may follow different annotation guidelines. Take \texttt{``\textcolor{darkgreen}{PERSON}''} for example, in MultiNERD~\cite{multiNERD_DATASET}, entities do not include the pronouns, e.g., ``he'' and ``she'', while ACE05~\cite{ACE2005_DATASET} consider personal pronouns as \texttt{``\textcolor{darkgreen}{PERSON}''}. To alleviate the problem, we add specific dataset information in the instructions to distinguish annotation guidelines for different datasets. For example, the instruction for the ACE05 dataset is ``... Please instantiate all the corresponding Event Objects in the following sentence \texttt{\textcolor{darkgreen}{from DATASET ACE05}}.''

\section{Experiment Setup}


\paragraph{Datasets.}
We conducted experiments using $33$ specific domain Information Extraction (IE) datasets, including $23$ datasets for Named Entity Extraction (NER), $8$ datasets for Relation Extraction (RE), $2$ datasets for Event Detection (ED) and Event Argument Extraction (EAE). The detailed statistics of these datasets are in Appendix \ref{appendix:stat}. Among these NER datasets, following ~\citet{wang2023instructuie, zhou2023universalner}, we take $7$ datasets as the zero-shot benchmark, including $5$ datasets of different domains from CrossNER~\cite{CrossNERDATASET}, MIT-Movie~\cite{MITReviewDataset} and MIT-Restaurant~\cite{MITReviewDataset}. For RE, we adopt GIDS~\cite{Jat2018ImprovingDS} as the zero-shot dataset. Following~\cite{sainz2023gollie}, we adopt CASIE~\cite{Lu2021Text2EventCS} as the zero-shot ED dataset.


To balance the evaluation coverage and costs, we introduce the KnowCoder benchmark, a composite derived from existing NER, RE, and EE datasets. Under the supervised setting, a sampling strategy was developed for NER and RE tasks to maintain the distributions of original datasets and ensure the broad coverage of knowledge types. Details on the proposed strategy and comprehensive benchmark information are available in Appendix~\ref{appendix:benchmark}. 

\begin{table*}
  \small
  \centering
  \setlength\tabcolsep{2pt}  
  \resizebox{0.8\linewidth}{!}
  {\begin{tabular}{@{}l|ccccccc|c@{}}
  \toprule
  \textbf{Model} &
  \textbf{Movie.} &
  \textbf{Rest.} &
  \textbf{AI} &
  \textbf{Litera.} &
  \textbf{Music} &
  \textbf{Politics} &
  \textbf{Science} &
  \textbf{Average} \\ \midrule

  LLaMA2-7B & 31.0 & 19.6 & 30.8 & 24.1 & 28.0 & 38.7 & 44.1 & 30.9 \\
  LLaMA2-13B  & 32.6 & 25.2 & 37.5 & 36.5 & 37.0 & 60.3 & \textbf{51.7} & 40.1 \\

  \midrule
  LLaMA2-7B & 31.0 & 19.6 & 30.8 & 24.1 & 28.0 & 38.7 & 44.1 & 30.9 \\
  KnowCoder-7B~(SU. only)   & \textbf{37.2}& \textbf{36.4} & \textbf{41.8} & \textbf{42.6} & \textbf{53.8} & \textbf{60.6} & 51.6 & \textbf{46.3}$^{\uparrow\textbf{49.8\%}}$ \\

  \bottomrule
  \end{tabular}}
  \caption{Results on NER under the few-shot setting.}
  \label{tab:fewshot-ner-results-table}
\end{table*}

\begin{table*}
  \centering
  \resizebox{1\linewidth}{!}
  {\begin{tabular}{@{}lcccccccc}
  \toprule
  \textbf{Model} &
  \textbf{Movie.} &
  \textbf{Rest.} &
  \textbf{AI} &
  \textbf{Litera.} &
  \textbf{Music} &
  \textbf{Politics} &
  \textbf{Science} &
  \textbf{Average} \\ 
  \midrule
  \textit{\textbf{w. refinement}} \\
  \rowcolor{softblue} InstructUIE-11B~\cite{wang2023instructuie} & - & - & 48.4 & 48.8 & 54.4 & 49.9 & 49.4 & -  \\
  \rowcolor{softblue} GoLLIE-7B~\cite{sainz2023gollie} & 63.0  & 43.4 & 59.1 & 62.7 & 67.8 & 57.2  & 55.5 & 58.4 \\
  \rowcolor{softblue} GoLLIE-13B~\cite{sainz2023gollie} & 62.5  & 49.8 & 56.7 & 59.7 & 65.5 & 54.4  & 56.2 & 57.8 \\
  \rowcolor{softblue} UniNER-7B~(refined)~\cite{zhou2023universalner} &  59.4 & 31.2 & 62.6 & 64.0 & 66.6 & 66.3 & 69.8 & 60.0 \\
  \midrule
  \textit{\textbf{w.o. refinement}} \\
  Vicuna-7B~\cite{chiang2023vicuna}  & 6.0 & 5.3 & 12.8 & 16.1 & 17.0 & 20.5 & 13.0 & 13.0  \\
  Vicuna-13B~\cite{chiang2023vicuna}  & 0.9 & 0.4 & 22.7 & 22.7 & 26.6 & 27.2 & 22.0 & 17.5  \\
  ChatGPT~\cite{ouyang2022training}  & 5.3 & 32.8 & 52.4 & 39.8 & 66.6 & 68.5 & \textbf{67.0} & 47.5  \\
  UniNER-7B~\cite{zhou2023universalner}  & 42.4 & 31.7 & 53.5 & 59.4 & 65.0 & 60.8 & 61.1 &53.4\\
  \rowcolor{softyellow} KnowCoder-7B   & \textbf{50.0} & \textbf{48.2} & \textbf{60.3} & \textbf{61.1} & \textbf{70.0} & \textbf{72.2} & 59.1 & \textbf{60.1}$^{\uparrow\textbf{12.5\%}}$\\
  \bottomrule
  \end{tabular}}
  \caption{Results on NER under the zero-shot setting. \hl{\textit{\textbf{w. refinement}}} denotes methods that are refined on human-annotated data, which is unfair for KnowCoder to compare with.}
  \label{tab:zeroshot-ner-results-table}
  \vspace{-4mm}
  \end{table*}

\paragraph{Metrics.}
We report the span-based offset Micro-F1 following previous methods~\cite{lu-etal-2022-unified, lin-etal-2020-joint}. For NER, an entity is considered correct if the entity boundary and type are correctly predicted. For RE, a relation is considered correct if its triplet matches a golden annotation, including relation type, subject entity, and object entity. For ED, an event trigger is correct if its event type and trigger match a golden annotation. For the EAE task, given an event type, an argument is correct if the argument and its role type match a golden annotation.

\subsection{Implementation Details}

KnowCoder is finetuned based on LLaMA2-base-7B~\cite{touvron2023llama2}. We
utilize the Megatron-LM framework~\cite{shoeybi2019megatron} for schema
understanding. We set the context length to $2048$, the learning rate to $5
\times 10^{-6}$, the global batch size to $1$M tokens, and the maximum training
step to $4500$. For the schema following and refinement phases, we use
LoRA~\cite{hu2021lora} for parameter-efficient fine-tuning. We set the lora rank
and lora alpha parameters to $32$ and $64$, respectively. The warmup ratio is
set to $0.03$ and the dropout ratio is set to $0.1$. The learning rates for
these two phases are set to $3 \times 10^{-4}$. We limit the sequence length to
$4096$ and set the batch size to $256$. Detailed information about the training
process is available in Appendix \ref{implementation_details}. During the
inference phase, we use greedy search and set the temperature to $0$. The
maximum output length is set to $640$.


\section{Results and Analyses}

\subsection{Few-shot Evaluation After Schema Understanding}

Considering that a pre-trained LLM cannot give proper results without given examples, we study the generalization ability of KnowCoder after the schema understanding phase, denoted as KnowCoder (SU. only), under the few-shot setting. Specifically, We utilize the first five samples from the training data as examples and report the NER F1 score in Table~\ref{tab:fewshot-ner-results-table} across seven zero-shot NER datasets. The results demonstrate that KnowCoder (SU. only) outperforms LLaMA2-7B with an average relative improvement of $\textbf{49.8\%}$. Remarkably, KnowCoder (SU. only) gets an average F1 score of $46.3\%$ with only a few examples, which are comparable to InstructUIE refined using human-annotated datasets. These results strongly support the effectiveness of the schema understanding phase in enhancing model generalization and performance in NER tasks.

\begin{table}
\centering
\small

\begin{tabular}{@{}r|r|l@{}}
\toprule

\textbf{Dataset}   & \multicolumn{1}{r|}{\textbf{SoTA}} & \multicolumn{1}{c}{\KnowCoder~\textbf{7B}} \\ \midrule
GIDS$_{RE}$      &     \cite{ouyang2022training} 9.9  &       \textbf{25.5}                        \\
CASIE$_{ED}$   &         \cite{sainz2023gollie} \textbf{59.3$^{\dag}$}          &       58.2                        \\
\midrule
Average    &             34.6                    &         \textbf{41.9$^{\uparrow\textbf{21.1\%}}$}                      \\ \bottomrule
\end{tabular}

\caption{Results on RE and ED tasks under the zero-shot setting. $^{\dag}$ indicates that it is unfair for KnowCoder to compare with the score.}
\label{tab:re-ed-zero-shot-results}
\vspace{-5mm}
\end{table}

\subsection{Zero-Shot Evaluation After Schema Following}

To verify the generalization ability of KnowCoder, we conduct zero-shot experiments on 9 datasets across NER, RE, and ED tasks. In this setting, we employ KnowCoder after schema understanding and following to conduct extraction. KnowCoder is compared with two kinds of baselines. One is the LLM-based IE method that refined on human-annotated data, including InstructUIE~\cite{wang2023instructuie}, GoLLIE~\cite{sainz2023gollie}, and UniNER~\cite{zhou2023universalner}. The other is models without refinement, including Vicuna~\cite{chiang2023vicuna}, ChatGPT, UniNER~\cite{zhou2023universalner}. The results of these three baselines are from \citet{zhou2023universalner}.  Note that KnowCoder is unfair when compared with methods after refinement.

\paragraph{Main Results.}
The results of zero-shot NER are in Table~\ref{tab:zeroshot-ner-results-table}.  It can be seen that KnowCoder surpasses baselines without refinement across four NER datasets, registering a relative performance enhancement of $\textbf{12.5\%}$. This improvement is attributed to KnowCoder's training on a large-scale, automatically generated dataset within a two-phase learning framework, which enhances its generalization capabilities for NER, even surpassing methods refined with human-annotated data. The results of zero-shot RE and ED are in Table~\ref{tab:re-ed-zero-shot-results}. For ED, KnowCoder's performance is inferior to GoLLIE, a baseline model trained on high-quality, human-annotated data. This emphasizes that human-annotated datasets can enhance performance for more difficult IE tasks, such as ED. To further substantiate the point, we further refine KnowCoder with the ACE05 dataset, the same EE training data employed by GoLLIE. This refinement significantly improves zero-shot F1 performance to $72.0\%$ on the CASIE dataset. This represents a significant advancement over GoLLIE's performance of $59.3\%$, marking a relative improvement of $\textbf{21.4\%}$.

\begin{table}
  \centering
  \setlength\tabcolsep{2.3pt}  
  \resizebox{1\linewidth}{!}
  {
  \begin{tabular}{@{}c|c|cccc|c@{}}
  \toprule
  \multirow{2}{*}{\textbf{Ratio}} & \multirow{2}{*}{\textbf{Model}} & \multicolumn{4}{c|}{\textbf{Task}} & \multirow{2}{*}{\textbf{Average}} \\
  \cmidrule{3-6}
  
  &       & ${NER}$   & ${RE}$   & ${ED}$  & ${EAE}$  &            \\ 
      \midrule
  \multirow{3}{*}{1\%}   & UIE-base                    &   \textbf{82.8}    &   30.8   &  41.5   &   12.8   &    42.0                  \\
  
  & LLaMA2-7B              &    72.3   &    32.1     &   35.3  &   33.3  &    43.3                  \\
         
  & KnowCoder-7B           &    79.2   &    \textbf{43.3}     &   \textbf{50.3}  &  \textbf{38.5}  &     \textbf{52.8$^{\uparrow\textbf{21.9\%}}$}                 \\ 
  
  \midrule
                         
  \multirow{3}{*}{5\%}   & UIE-base                   &   88.3    &    \textbf{51.7}   &  55.7   &   30.4  &            56.5            \\
  
  & LLaMA2-7B              &   89.3    &    35.7     &   52.6  &   46.3  &    56.0                \\
  
  & KnowCoder-7B           &   \textbf{90.6}    &    51.1     &   \textbf{59.0}  &   \textbf{48.3} &      \textbf{62.3$^{\uparrow\textbf{10.3\%}}$}               \\ 
     
  \midrule
  
  \multirow{3}{*}{10\%}  & UIE-base                 &   89.6    &   \textbf{59.2}   &   60.3  &    36.3  &          61.4            \\
  
  & LLaMA2-7B              &   91.2    &    48.6     &   60.7  &   52.3 &      63.2                \\
  
  & KnowCoder-7B           &   \textbf{92.2}    &    53.6     &   \textbf{62.2}  &   \textbf{55.1} &      \textbf{65.8$^{\uparrow\textbf{4.1\%}}$}                \\ 
  
  \bottomrule
  \end{tabular}
  }
  \caption{
      Low-resource results on IE tasks, where \textbf{Average} is the average F1 across four IE tasks.
  }
  \label{tab:lowresource}
  \vspace{-3mm}
\end{table}

\subsection{Low Resource Evaluation After Schema Following} 

To further investigate the generalization ability of KnowCoder for IE tasks, we conduct low-resource experiments by fine-tuning KnowCoder with three different partitions of the original training sets (1/5/10\% ratio) across four tasks. Following \citet{lu-etal-2022-unified},
we adopt CoNLL03, CoNLL04, ACE05$_{ED}$ and ACE05$_{EAE}$ as the benchmarks for NER, RE, ED, and EAE tasks. LLaMA2 denotes directly fine-tuning LLaMA2 with these partial training data. The results are in Table~\ref{tab:lowresource}. It
can be shown that KnowCoder gets the highest average F1 scores across all IE tasks in low-resource settings at varying ratios. In ratio $1\%$, KnowCoder gets the relative average improvement of $\textbf{21.9\%}$ compared to UIE, which shows that KnowCoder has strong adaptability to downstream IE tasks after pretraining on large-scale data under the two-phase learning framework.

\subsection{Supervised Evaluation After Refinement}

Under the supervised evaluation, KnowCoder is further refined with the IE datasets. We conduct supervised experiments on four IE tasks, including NER, RE, ED, and EAE. KnowCoder is compared with three kinds of methods. The first is the traditional UIE method~\cite{lou2023universal, lu-etal-2022-unified}, which is based on relatively small language models (i.e., million-level parameters). The latter two are based on LLMs (i.e., ChatGPT, LLaMA2). They adopt the in-context learning~\cite{guo2023retrieval, codeie, ashok2023promptner} and supervised fine-tuning paradigms~\cite{zhou2023universalner,wang2023instructuie,sainz2023gollie}, respectively. As some baselines only report results for specific IE tasks, we report the SOTA results of the above methods in each dataset, denoted as ``SoTA'' in the tables. As highlighted by ~\citet{zhou2023universalner}, the evaluation script of InstructUIE~\cite{wang2023instructuie} contains issues. Furthermore, the benchmark in ~\citet{zhou2023universalner} remains pending release. In the end, we have implemented these two baselines on KnowCoder benchmark using their released models.

\begin{table}[]
  \centering
  \setlength\tabcolsep{1.6pt}  
  \resizebox{0.95\linewidth}{!}{
  \begin{tabular}{@{}c|r|c@{}}
  \toprule
  
  \textbf{Dataset}   & \multicolumn{1}{r|}{\textbf{SoTA}} & \multicolumn{1}{c}{\KnowCoder~7B} \\ \midrule
  ACE04      &         \cite{lu-etal-2022-unified} \textbf{87.6}               &      86.2                        \\
  ACE05      &         \cite{sainz2023gollie} \textbf{89.6}               &       86.1                        \\
  AnatEM   &         \cite{zhou2023universalner} \textbf{88.9}         &       86.4                       \\
  Broad Twitter  &         \cite{zhou2023universalner} \textbf{79.8}                  &     78.3                        \\
  CoNLL03   &         \cite{zhou2023universalner} 94.8          &       \textbf{95.1}                        \\
  DIANN      &          \cite{sainz2023gollie} 84.1         &       \textbf{94.7}                       \\
  FabNER &          \cite{zhou2023universalner} 82.3        &       \textbf{82.9}                      \\
  FindVehicle        &         \cite{zhou2023universalner} 98.4       &       \textbf{99.4}                     \\
  GENIA        &         \cite{zhou2023universalner} \textbf{80.3}        &       76.7                    \\
  Movie       &         \cite{zhou2023universalner} 90.2        &       \textbf{90.6}                        \\
  Rest. &          \cite{wang2023instructuie} \textbf{82.6}        &       81.3                    \\
  MultiNERD       &         \cite{zhou2023universalner} 93.9        &       \textbf{96.1}                        \\
  OntoNotes 5  &         \cite{sainz2023gollie} 84.6                  &      \textbf{88.2}                        \\
  WikiANN     &         \cite{zhou2023universalner} 85.4                  &      \textbf{87.0}                       \\ 
  WNUT17     &         \cite{sainz2023gollie} 54.3                  &      \textbf{66.4}                       \\ 
  bc2gm      &          \cite{wang2023instructuie} 80.5         &       \textbf{82.0}                      \\
  bc5cdr     &         \cite{zhou2023universalner} \textbf{91.5}                  &       89.3                        \\ 
  ncbi     &         \cite{wang2023instructuie} \textbf{85.0}                  &  83.8                     \\
  \midrule
  Average    &             85.2                    &         \textbf{86.1$^{\uparrow\textbf{1.1\%}}$}                      \\ \bottomrule
  \end{tabular}}
  \caption{Results on NER under the supervised setting.}
  \vspace{-4mm}
  \label{tab:ner-supervised-results}
\end{table}

\paragraph{Main Results.}
The results for NER, RE, EE (including ED and EAE) tasks are shown in Tables~\ref{tab:ner-supervised-results}, ~\ref{tab:re-supervised-results} and ~\ref{tab:ed-eae-supervised-results}, respectively. We can observe that: (1) KnowCoder outperforms the SOTA baselines on most datasets for NER, RE, ED, and EAE, respectively. Based on the code-style schemas, KnowCoder universally models IE tasks and effectively transfers IE abilities after conducting schema understanding, following, and refinement on large-scale training data. (2) In more challenging UIE tasks, such as RE, KnowCoder demonstrates impressive advancements up to the relative improvement of 8.6\% compared to the SOTA baselines. KnowCoder achieves the performances of 73.9\% for ED and 66\% for EAE. This is \textbf{the first time} LLM-based UIE methods surpass smaller models like UIE in ED and EAE tasks. The code-style schemas and the learning framework enable a more precise definition and understanding of this complex structured knowledge, leading to a significant improvement. (4) UniNER~\cite{zhou2023universalner} achieves comparable results to KnowCoder on NER. Nonetheless, KnowCoder surpasses UniNER in several respects. Primarily, UniNER is limited to extracting one type of entity per iteration, leading to a cost-time complexity. In contrast, KnowCoder can extract multiple entity types in a single iteration, enhancing efficiency. Additionally, UniNER relies on a text-style schema, making it hard to represent and extract relations and events effectively. Conversely, KnowCoder, as a UIE model, offers broader versatility and efficacy comparing to UniNER.  (3) KnowCoder gets better results than baselines with code-style prompt ~\cite{codeie, guo2023retrieval, sainz2023gollie}. This is because KnowCoder provides a more comprehensive schema representation method and conducts two-phase training to understand and follow these schemas.

\begin{table}[]
  \centering
  \setlength\tabcolsep{2pt}  
  \resizebox{0.85\linewidth}{!}{
  \begin{tabular}{@{}c|r|c@{}}
  \toprule
  
  \textbf{Dataset} & \textbf{SoTA} & \KnowCoder~7B \\
  \midrule
  ACE05      &        \cite{sainz2023gollie}~\textbf{70.1}               &       64.5                        \\
  semevalRE   &       \cite{wang2023instructuie}~65.8          &       \textbf{66.3}                        \\
  CoNLL04  &          \cite{lou2023universal}~\textbf{78.8}                  &       73.3                        \\
  NYT        &        \cite{wang2023instructuie}~91.0        &       \textbf{93.7}                       \\
  ADE corpus &        \cite{wang2023instructuie}~82.8         &       \textbf{84.3}                        \\
  kbp37      &        \cite{wang2023instructuie}~30.6         &       \textbf{73.2}                        \\
  GIDS      &         \cite{wang2023instructuie}~76.9         &       \textbf{78.0}                        \\
  SciERC     &        \cite{lou2023universal}~37.4                  &       \textbf{40.0}                        \\ \midrule
  Average    &        66.7                        &         \textbf{71.7$^{\uparrow\textbf{7.5\%}}$}                      \\ \bottomrule
  \end{tabular}}
  \caption{Results on RE under the supervised setting.}
  \label{tab:re-supervised-results}
\end{table}

\begin{table}
  \centering
  \resizebox{0.8\linewidth}{!}
  {
  \begin{tabular}{@{}c|cc}
  \toprule
  \textbf{Model}  & \textbf{ACE05$_{ED}$}         & \textbf{ACE05$_{EAE}$}    \\ 
  
  \midrule
  UIE                     &   73.4    &   69.3   \\
  USM                     &   69.3    &   63.3   \\ 
  \midrule
  Code4UIE                &   37.4    &   57.0   \\ 
  
  \midrule
  
  InstructUIE-11B             &   43.2    &   56.8   \\
  GoLLIE-7B                  &   72.2    &   66.0   \\ 
  
  \midrule
  KnowCoder-7B              &   \textbf{74.2}   &   \textbf{70.3}     \\
  \bottomrule
  \end{tabular}}
  \caption{Results on ED and EAE under the supervised setting.}
  \label{tab:ed-eae-supervised-results}
  \vspace{-4mm}
\end{table}

\subsection{Ablation Study}
To show how the schema following and understanding phases contribute to KnowCoder under the zero-shot setting, we further conduct ablation studies removing the schema understanding and following phase, denoted as KnowCoder (w.o. SU) and KnowCoder (w.o. SF), respectively. The results on seven zero-shot NER datasets are shown in Table~\ref{tab:ablation_zero_shot}. It can be seen that: (1) KnowCoder gets better results than KnowCoder (w.o. SF) on most NER datasets. It is because the schema understanding phase helps KnowCoder to understand concepts in the schema by training on definition and instance codes and increases its generalization ability. (2) Results of KnowCoder (w.o. SF) decrease extremely, which proves the importance of schema following. Due to the lack of in-context learning ability, a 7B model without instruction tuning is hard to understand instructions under the zero-shot setting, thus making it hard to finish the IE tasks.

\section{Related Work}
\subsection{Universal Information Extraction}
Universal information extraction aims to conduct different IE tasks via a single
model. The existing UIE models first represent different schemas for IE tasks in
a universal way. OneIE~\cite{lin2020joint} represents schemas as classification
labels, InstructUIE~\cite{wang2023instructuie} uses
keywords~\cite{gui2023instructie, lou2023universal} of concepts to represent
schemas, and UIE~\cite{lu-etal-2022-unified} uses a specifically-designed formal
language to represent schemas. Based on such schema representations, these
models adopt language models to understand the schemas and extract the
corresponding structured knowledge. 

\begin{table}[]
\centering
\resizebox{0.85\linewidth}{!}
{
\begin{tabular}{@{}c|c|cc@{}}
\toprule
\textbf{Dataset}  & \KnowCoder~\textbf{7B} & \textbf{w.o. SU} & \textbf{w.o. SF} \\ 

\midrule

Movie.   & 50.0      & +1.6    & \textbf{-50.0}      \\
Rest.    & 48.2      & \textbf{-0.8}    & \textbf{-46.1}      \\
AI       & 60.3      & \textbf{-4.5}    & \textbf{-57.7}      \\
Litera.  & 61.1      & +0.6    & \textbf{-59.0}      \\
Music    & 70.0      & \textbf{-3.1}    & \textbf{-69.0}      \\
Politics & 72.2      & \textbf{-1.8}    & \textbf{-70.8}      \\
Science  & 59.1      & \textbf{-2.7}    & \textbf{-55.6}      \\


\bottomrule
\end{tabular}}
\caption{Ablation study under the zero-shot setting.}
\label{tab:ablation_zero_shot}
\vspace{-4mm}
\end{table}

\subsection{Large Language Models for IE}
Due to the strong generation abilities of LLMs, they have been used in
IE recently~\cite{xu2023large}. LLM-based IE methods can be divided into
two categories: In-Context Learning (ICL) based methods and Supervised Finetuning (SFT) based methods. The
ICL-based IE methods~\cite{codeie, guo2023retrieval, ashok2023promptner,
wang2023gpt} make predictions only based on contexts augmented with a few
examples. The SFT-based methods~\cite{wangdeepstruct, gui2023instructie,
wang2023instructuie, zhou2023universalner, xu2023large, sainz2023gollie} use the
annotated data to finetune LLMs.

Some existing work uses code-style prompts to conduct IE. Most of them are
ICL-based methods. ~\citet{wang2022code4struct} uses the code-style prompt to
conduct event argument extraction. ~\citet{codeie} uses the code-style prompt to
conduct the named entity extraction and relation extraction.
~\cite{guo2023retrieval} proposes a reterive-argumented method to conduct the
universal IE. These methods show relatively poor performance compared to
SFT-based methods because of the lack of training to follow the schemas in the
prompt. The most similar work with KnowCoder is GoLLIE, an SFT-based UIE method
that gives out definitions of schemas as code comments. The difference between
KnowCoder and GoLLIE is that KnowCoder designs a more comprehensive code-style
schema representation method, including taxonomies, constraints, and class
methods, and further constructs a large-scale schema library. Besides, GoLLIE
conducts instruction tuning on human-annotated data, while KnowCoder contains a
two-phase learning framework that enhances schema understanding and following
ability via automatically annotated data. 

\section*{Conclusion}

  In this paper, we introduced KnowCoder for UIE leveraging Large Language
  Models. KnowCoder is based on a code-style schema representation method and
  an effective two-phase learning framework. The code-style schema representation
  method uniformly transforms different schemas into Python classes, with which
  the UIE task can be converted to a code generation process. Based on the schema representation
  method, we constructed a comprehensive code-style schema library covering over $30,000$
  types of knowledge. To let LLMs understand and follow these schemas, we
  further proposed a two-phase learning framework that first enhances the
  schema comprehension ability and then boosts its schema following ability. After training on billions of automatically annotated data and refining with human-annotated IE datasets, KnowCoder demonstrates remarkable performance improvements on different IE tasks under the various evalution settings.

\section*{Limitations}
The schemas utilized in our approach are predominantly constructed from Wikidata, which occasionally results in some schemas lacking definitions or other relevant information. This necessitates the generation of additional data to supplement these missing elements. During the pretraining phase, we adopted a combination of automatic generation and distant supervision methods to amass a large corpus. However, this approach inevitably introduces a certain degree of noise. Furthermore, there remains room for improvement in terms of the richness and complexity of the current corpus. Further exploration of pretraining settings could also be beneficial in enhancing the zero-shot capabilities for relation and event-related tasks.

\bibliography{acl_latex}

\appendix


\label{sec:appendix}

\begin{table*}[htbp]

\centering
\resizebox{0.85\linewidth}{!}{
\begin{tabularx}{\textwidth}{c|p{0.2\textwidth}|X|X}
\toprule
\multirow{2}{*}{\textbf{Class Name}} & \multirow{2}{*}{\textbf{Class Method}} & \multicolumn{2}{c}{\textbf{Results}} \\
\cline{3-4}
 &  & \textbf{w.o. Class Method} & \textbf{w. Class Method} \\ 

\midrule

Average Ratings  & If the extracted span is a number, add ``star'' after the number if ``star'' follows the number in the sentence. & I am looking for a unrated disney movie about a teddy bear starring julie pinson with a \hl{four} star ratings average. & I am looking for a unrated disney movie about a teddy bear starring julie pinson with a \hl{four star} ratings average. \\ 

\midrule

Facility  & Delete the content after the word ``as'' if ``as'' in the extracted span. & It lies just 12 miles from Baghdad and will be \hl{a key forward base for U.S. troops as they prepare for a push on the capital}. & It lies just 12 miles from Baghdad and will be \hl{a key forward base for U.S. troops} as they prepare for a push on the capital.  \\ 

\midrule

Organization  & Delete the content after the word ``such as'' if ``such as'' in the span. & Megawati and Putin are expected to sign agreements to give Russian companies a toehold in Indonesia's oil and gas industry, long dominated by \hl{American and British giants such as Exxon Mobil and BP}. & Megawati and Putin are expected to sign agreements to give Russian companies a toehold in Indonesia's oil and gas industry, long dominated by \hl{American and British giants} such as Exxon Mobil and BP. \\

\bottomrule
\end{tabularx}
}
  \caption{Cases of class methods.}
    \vspace{-1mm}
  \label{tab:class_methods}
\end{table*}


\section{Analyses on Schema Importing}\label{sec:import_position}
To get insight into how the data organization method contributes to the results of schema understanding, we compare the performance of KnowCoder training on different version data with different schema importing methods. Specifically, we generate three versions of training data, named ``Import-First'',
``Sentence-First'', and ``Whole''. ``Import-First'' denotes
that we import the class first and then give out the sentence.
``Sentence-First'' denotes that we import the class following the
sentence, which is the version KnowCoder adopts. ``Whole'' denotes the
version that we give out the class after the sentence with their whole
definitions. We train the model under the same setting and report the micro F1
curve on the test set of seven zero-shot NER datasets. The results are shown in Figure ~\ref{fig:F1-curve-under-different training
data}. It can be seen that ``Sentence-First'' performs best. If the
``import'' clause is before the sentence, LLMs are trained to predict the
specific class after \texttt{``\textcolor{darkgreen}{from Entities import}''} without giving any information.
The ``Whole'' method makes the model overfitting to the definition code
because they are repeated frequently.

\section{Analyses on Class Name}

\begin{figure}[tbp]  
  \centering
  \includegraphics[width=0.45\textwidth]{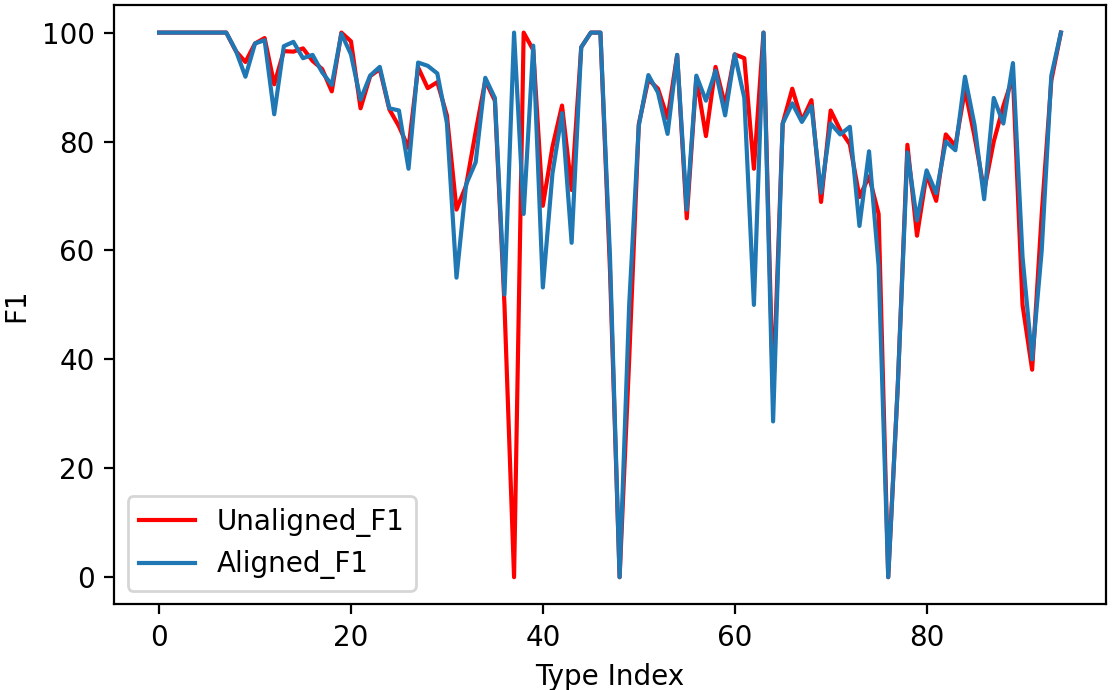}
  \caption{F1 scores of KnowCoder on each type before and after alignment for NER task.}
 \vspace{-4mm}
  \label{fig:schema_ali}
\end{figure}

The same concept may have different names in different IE datasets, and the concept name in downstream datasets may conflict with the name in KnowCoder schema. For example, \texttt{``\textcolor{darkgreen}{Human}''} in KnowCoder schema shares the same meaning as \texttt{``\textcolor{darkgreen}{Person}''} in ACE05. To eliminate conflicts among names of concepts in different schemas, we align the concept names in IE datasets to KnowCoder schema. Note that, for a fair comparison, we make sure the number of concepts in a dataset does not change during the alignment process. Figure \ref{fig:schema_ali} illustrates the F1 performance across all types under aligned and unaligned experimental settings. With average scores of $81.37$ and $81.35$, respectively, it can be inferred that aligning schemas does not significantly impact the model's outcomes.

\begin{figure}[tbp]  
  \centering
  \includegraphics[width=0.45\textwidth]{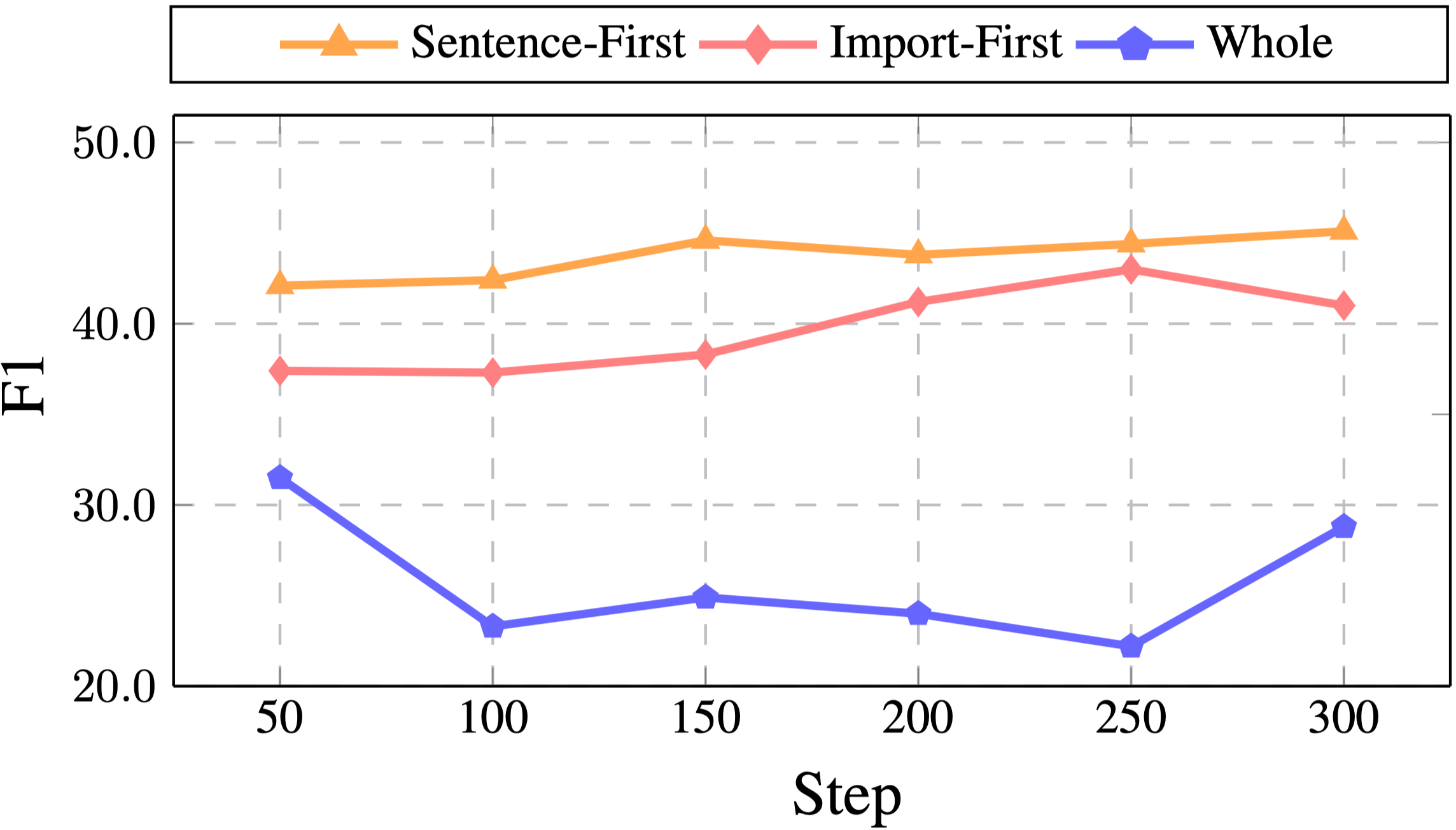}
  \caption{Detailed Analysis of different schema importing methods.}
  \label{fig:F1-curve-under-different training data}
\end{figure}

\begin{table*}
\centering
\setlength\tabcolsep{2pt}  
{
\begin{tabular}{@{}c|ccccccccc@{}}
\toprule
\textbf{Model} &
\textbf{ACE05} &
\textbf{Bro. Twi.} &
\textbf{Movie.} &
\textbf{Rest.} &
\textbf{Ncbi.} &
\textbf{Ave} \\ \midrule

KnowCoder-7B   & 85.0& 77.9 & 90.6 & 81.3 & 82.8 &  83.5 \\
+ Class Methods  & $\uparrow\textbf{0.9}$ & $\uparrow\textbf{1.1}$ & $\uparrow\textbf{0.9}$ & $\uparrow\textbf{1.2}$ & $\uparrow\textbf{0.7}$  & $\uparrow\textbf{1.0}$ \\


\bottomrule
\end{tabular}}
\caption{Results on IE tasks with Class Methods, where \textbf{Ave} is the average F1 across five datasets.}
\label{tab:class-method-results-table}
\vspace{-4mm}
\end{table*}

\section{Analyses on Class Methods}\label{sec:class_method}

Class methods are utilized to post-process the extracted results generated by LLMs. Three cases of the used class methods are listed in Table~\ref{tab:class_methods}. To demonstrate the effectiveness of class methods, We conduct experiments on five NER datasets, including ACE05, Broad Twitter, MIT Movie, MIT Restaurant, and Ncbi-disease. The results are shown in Table~\ref{tab:class-method-results-table}. It can be observed that KnowCoder gets an average F1 improvement of $1\%$. By defining some class-specific extraction rules, class methods help KnowCoder to extract more precise results.

\section{Analyses on Prompts}\label{sec:prompt}
To validate the influence of different prompts on the results, Table~\ref{tab:prompt_template} reports the performance of NER on ACE05 using prompts with two styles, i.e., Code and IE styles. It can be observed that results are similar (with a gap of $0.7\%$ of F1), which verifies the robustness of KnowCoder to different prompts. The code-style prompt is slightly better than the IE style, suggesting that code-style prompts can better stimulate the code generation capabilities of LLMs compared to the text-style prompt and thus benefit the IE tasks.

\begin{table*}
\centering
\begin{tabularx}{\textwidth}{c|c|>{\centering\arraybackslash}p{0.75\textwidth}|c}
\toprule
\textbf{No.} & \textbf{Style} & \textbf{Template} & \textbf{F1} \\ 

\midrule


1 & Code & Some Classes are defined above. Please instantiate the Objects corresponding to the above Classes in the sentence. & 82.4 \\ 

\midrule

2 & IE & Some Entity Types are given above. Please find all the Entities in the above Types in the sentence.  & 81.6 \\ 

\bottomrule
\end{tabularx}
  \caption{Performance of prompts with different styles on the NER task.}
  \label{tab:prompt_template}
\end{table*}

\section{Analyses on Negative Sampling}

To demonstrate how the negative class sampling and fully negative sample
construction contribute to the results, we conduct experiments of removing the
negative classes~(denoted as w.o. NC) and fully negative samples~(denoted as
w.o. FNS), respectively. The macro average of F1 on seven zero-shot NER datasets is reported
in Table~\ref{tab:ns-results-table}. It can be seen that the performance of KnowCoder decreases
without negative sampling, which proves the effectiveness of the negative class
sampling and fully negative sample construction.

\begin{table}
  \centering
  {\begin{tabular}{l|ccc}
  \toprule
  \textbf{Model} &
    \textbf{\KnowCoder~7B} &
    \textbf{w.o. NC} &
    \textbf{w.o. FNS} \\
  \midrule
  zero-shot F1 & \textbf{57.8} & 50.4 $^{\downarrow\textbf{7.4}}$ & 55.7$^{\downarrow\textbf{2.1}}$\\
  \bottomrule
  \end{tabular}}
  \caption{Detailed Analysis of the Negative Sampling.}
  \label{tab:ns-results-table}
\end{table}

\section{KnowCoder Benchmark}\label{appendix:benchmark}

\paragraph{Benchmark Construction.}
Considering the significant expenses associated with assessing all test sets for NER and RE tasks, we developed a sampling method to establish the KnowCoder Benchmark to balance evaluation expenses and precision. Our primary principle is ensuring the sampled subset retains the same distribution. Specifically, we randomly sampled a portion of samples from each type in the dataset with a scaling factor $s$. For NER and RE tasks, we set $s$ to $14$ and $4$, respectively. Assuming the original number of samples of the type in a dataset is $x$, the sampled number in the benchmark is:
\begin{eqnarray}
k=\lceil x/s \rceil, s\geq1.
\label{eq:sample_num}
\end{eqnarray}

Note that we adopted the same sampling method for the empty samples in datasets. Moreover, a sample may be sampled multiple times because there may be more than one type of instance. Thus, we remove duplicate samples during the sampling process. Due to the smaller number and size of EAE and ED datasets, we used the complete dataset for evaluation.

\paragraph*{Statistics of the Benchmark.}

Table \ref{tab:benchmark_stat} summarizes the information on the benchmarks under the supervised setting for two tasks: NER and RE.

\begin{table}[H]
\centering
\begin{tabular}{@{}l|ccc@{}}
\toprule
\textbf{Task} & \textbf{\#Sample} & \textbf{\#Type} & \textbf{\#Source} \\ \midrule
NER           & 8287              & 92              & 18                \\
RE            & 5009              & 64              & 8                 \\ \bottomrule
\end{tabular}
\caption{Statistics of the benchmark build on NER and RE tasks under the supervised setting.}
\label{tab:benchmark_stat}
\end{table}

\paragraph{Benchmark Significance.}
The results reported in this paper are produced in the sampled benchmark with $42$ as the base seed. To systematically assess how the generated benchmarks affect the reproducibility and consistency of the model's effectiveness, we employ multiple rounds of experiments on benchmarks with distinct random seeds, i.e., $1$, $2$, and $42$. Table \ref{tab:benchmark_significance} summarizes the average performance on NER and RE tasks. It can be observed that the performance variations of KnowCoder across different benchmarks are minor ($85.1\pm{0.2}$ for NER and $71.9\pm{0.5}$ for RE). The results demonstrate that KnowCoder's results reported in this paper are both consistent and reproducible.

\begin{table}
\centering
\begin{tabular}{@{}r|cc}
\toprule
\textbf{Base Seed} & \textbf{NER} & \textbf{RE} \\ \midrule
1                  & 85.0        & 72.5       \\
2                  & 84.9        & 71.7       \\
42                 & 85.3        & 71.6       \\ \bottomrule
\end{tabular}
\caption{Results of NER and RE tasks on benchmarks with different random seeds.}
\label{tab:benchmark_significance}
\end{table}

\section{Training Data Generation}\label{sec: data_clean}
The training data used in the schema understanding phase consists of two kinds of codes, i.e., schema definition codes and instance codes. In this section, we will give more details of the instance code generation process.

The instance code is generated based on the KELM corpus.
The processing procedure mainly includes four steps: entity typing, entity code generation, relation code generation, and event code generation. The origin data of the KELM corpus does not annotate the types of entities. We obtain the mappings from entity names to entity types based on WikiData. Specifically, we find the corresponding WikiData ID for each entity in KELM and identify its types through the \texttt{``\textcolor{darkgreen}{InstanceOf}''} relations. For those entities without types, we filter them from training data. Then, we generate the entity code based on the typed KELM corpus. Finally, we clean the data by removing samples with the entity type ``Wikimedia Disambiguation Page'' and removing contents in brackets for entities and entity types. Based on the typed entities, we generate the relation code. Since KELM does not contain event codes, we consider relations to be events if they have sub-properties. We treat their relation types as event types, the sub-properties as corresponding role types, and the annotated mentions as arguments. Furthermore, we delete samples if the event role is one of \textit{``of''}, \textit{``follows''}, \textit{``followed by''}, \textit{``point in
time''}, \textit{``country''}.

\begin{table*}[]
\centering

\resizebox{1\linewidth}{!}
{\begin{tabular}{@{}c|cc|ccccc}
\toprule
\textbf{Phase}                                 & \textbf{Task} & \textbf{Data Name}    & \textbf{\#Types} & \textbf{\#Instance} & \textbf{\#Tokens} & \multicolumn{1}{c}{\textbf{Disk size}} & \multicolumn{1}{c}{\textbf{Hierarchy}} \\ \midrule
\multirow{3}{*}{Schema Understanding} & NER  & KELM         & 19,009  & 2,019,990  & 0.26B    &     1.15GB                    &      \greencheck                         \\
                                      & RE   & KELM         & 810     & 1,191,199  & 0.13B    &     0.54GB                   &        \greencheck             \\
                                      & EE   & KELM         & 499     & 296,403    & 0.03B    &    0.11GB                    &                  \redmark             \\ \midrule
\multirow{4}{*}{Schema Following}     & NER  & UniversalNER & 12,072  & 127,839    & 0.19B    &    0.96GB                    &        \greencheck             \\
                                      & RE   & InstructIE   & 131     & 327,984    & 0.62B    &      2.61GB                   &        \greencheck             \\
                                      & ED   & LSEE         & 20      & 415,353    & 0.26B    &      1.03GB                   &         \redmark             \\
                                      & EAE  & LSEE         & 20      & 211,635    & 0.10B    &      0.50GB                   &          \redmark             \\ \bottomrule
\end{tabular}}
\caption{Statistics of schema understanding instance codes and schema following instruction tuning codes.}
\label{schema_understanding}
\end{table*}

\section{Data Statistics}

\label{appendix:stat}
\paragraph*{Statistics of the Constructed Schema Library.}

The schema library is constructed on
KELM~\cite{agarwal-etal-2021-knowledge},
UniversalNER~\cite{zhou2023universalner}, InstructIE~\cite{knowlm} and LSEE~\cite{chen-etal-2017-automatically}. The detailed analysis
of each task schema is shown in Table \ref{schema_stat}. Here, ``\#Type'' denotes
the total number of types, ``\#Type w/ desc.'' indicates the count of types with
descriptions, and ``\#Type w/o desc.'' signifies the count of types without
descriptions.

\begin{table}[h]
\centering
\resizebox{1\linewidth}{!}
{\begin{tabular}{cccc}
\toprule
\textbf{Task} & \textbf{\#Type} & \textbf{\#Type w/ desc.} & \textbf{\#Type w/o desc.} \\ \midrule
NER & 29,177 & 19,856 & 9,321 \\ 
RE & 876 & 840 & 36 \\ 
EE & 519 & 515 & 4 \\ 
\bottomrule
\end{tabular}}
\caption{Statistics of the constructed schema library.}
\label{schema_stat}
\end{table}

\paragraph*{Statistics of the Training Data.}

The training data consists of three parts: schema understanding data, schema following data, and specific domain IE data. The schema understanding training data includes
schema definition codes and instance codes. The schema definition codes are built based on the schema
library, with statistical results shown in Table \ref{schema_stat}. Schema instance codes are constructed based on
KELM~\cite{agarwal-etal-2021-knowledge}, with statistical results provided in
Table \ref{schema_understanding}. The schema following training data is constructed on UniversalNER~\cite{zhou2023universalner}, InstructIE~\cite{knowlm} and LSEE~\cite{chen-etal-2017-automatically}. The statistics of
schema following training data are presented in Table \ref{schema_understanding}.

Additionally, for specific domain Information Extraction (IE), we conduct experiments utilizing $33$ datasets, comprising $23$ datasets for the NER task, $8$ datasets for the RE task, and $2$ datasets for the ED and EAE tasks. Specifically, under the supervised setting, we employ $18$ datasets for the NER task, including ACE04~\cite{mitchell2005ace}, ACE 2005~\cite{ACE2005_DATASET}, AnatEM~\cite{AnatEM_DATASET}, Broad
Twitter~\cite{broad_twitter_corpus_DATASET},
bc2gm~\cite{Kocaman2020BiomedicalNE_DATASET},
bc5cdr~\cite{Li2016BioCreativeVC_DATASET}, 
CoNLL03~\cite{CoNLL03_Dataset},
DIANN~\cite{pan-etal-2017-cross}, 
FabNER\cite{Kumar2021FabNERIE}, 
FindVehicle~\cite{FindVehicle_DATASET},
GENIA~\cite{GENIANER_DATASET}, 
MIT Movie~\cite{MITReviewDataset}
MIT Restaurant~\cite{MITReviewDataset}
MultiNERD~\cite{multiNERD_DATASET}, 
ncbi-disease~\cite{ncbi-disease_DATASET},
Ontonotes5~\cite{weischedel2013ontonotes},
WikiANN~\cite{wikiann_Dataset},
and WNUT17~\cite{derczynski2017results}. For the RE task, we utilize $8$ datasets under the supervised setting, including ACE
2005~\cite{ACE2005_DATASET}, 
ADE corpus~\cite{ADEcorpus_DATASET}, 
CoNLL04~\cite{Roth2004ALP},
GIDS~\cite{Jat2018ImprovingDS},
kbp37~\cite{kbp37_DATASET}, 
NYT~\cite{NYT_DATASET},
SciERC~\cite{SciERC_DATASET}, 
and semeval RE~\cite{Hendrickx2010SemEval2010T8}. For the ED and EAE tasks, ACE05~\cite{ACE2005_DATASET} and CASIE~\cite{Lu2021Text2EventCS} are employed.

Under the zero-shot setting, we take $7$ datasets for the NER task, following ~\citet{wang2023instructuie, zhou2023universalner}, which include $5$ CrossNER subsets (AI, literature, music, politics, science)~\cite{CrossNERDATASET}, MIT-Movie~\cite{MITReviewDataset} and MIT-Restaurant~\cite{MITReviewDataset}. For the RE task, we adopt GIDS~\cite{Jat2018ImprovingDS} under the zero-shot setting. For the ED and EAE tasks, CASIE~\cite{Lu2021Text2EventCS} is adopted under the zero-shot setting, following~\cite{sainz2023gollie}.

\begin{figure}[t]
\small
\centering
  \includegraphics[width=1.0\linewidth]{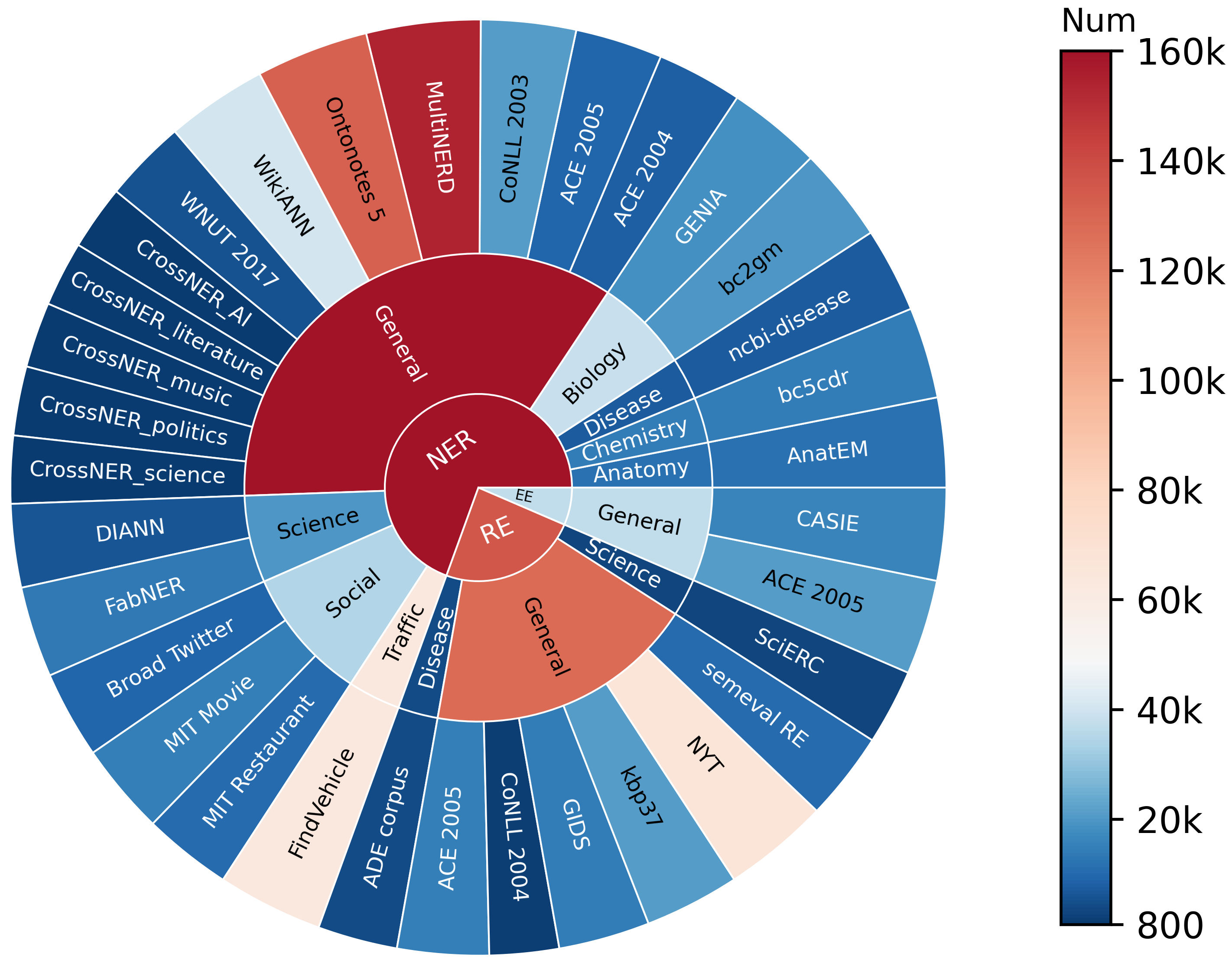}
  \caption{Overview of the datasets on specific domain IE.}
 \label{data_stat_pie}
\end{figure}

The detailed statistic of each dataset is shown in Table \ref{sft2-stat}. Here, ``\#Type'' indicates the number
of types, while ``\#Train'', ``\#Dev'', and ``\#Test'' denote the number of
sentences in the training, development, and test datasets, respectively. Figure~\ref{data_stat_pie} shows the overview of the datasets on specific domain IE by task and size. Note that the statistics for each dataset in the figure encompass the total number of train, dev, and test datasets.

\begin{table}[htbp]
    \centering
    \resizebox{1\linewidth}{!}{
    \begin{tabular}{c|c|c|ccc}
    \toprule  
    \textbf{Task} & \textbf{Dataset} & \textbf{\#Type} & \textbf{\#Train} & \textbf{\#Dev} & \textbf{\#Test} \\
    \midrule
    \multirow{23}*{NER} & ACE04 & 7 & 6,202 & 745 & 812 \\
    ~ & ACE05 & 7 & 7,299 & 971 & 1,060 \\
    ~ & AnatEM & 1 & 5,861 & 2,118 & 3,830 \\
    ~ & bc2gm & 1 & 12,500 & 2,500 & 5,000 \\
    ~ & bc5cdr & 2 & 4,560 & 4,581 & 4,797 \\
    ~ & Broad Twitter & 3 & 5,334 & 2,001 & 2,000 \\
    ~ & CoNLL03 & 4 & 14,041 & 3,250 & 3,453 \\
    ~ & DIANN & 1 & 3,900 & 975 & 1,334 \\
    ~ & FabNER & 12 & 9,435 & 2,182 & 2,064 \\
    ~ & FindVehicle & 21 & 21,565 & 20,777 & 20,777 \\
    ~ & GENIA & 5 & 15,023 & 1,669 & 1,854 \\
    ~ & MIT Movie & 12 & 9,774 & 2,442 & 2,442 \\
    ~ & MIT Restaurant & 8 & 7,659 & 1,520 & 1,520 \\
    ~ & MultiNERD & 16 & 134,144 & 10,000 & 10,000 \\
    ~ & ncbi-disease & 1 & 5,432 & 923 & 940 \\
    ~ & Ontonotes 5 & 18 & 107,032 & 14,110 & 10,838 \\
    ~ & WikiANN & 3 & 20,000 & 10,000 & 10,000 \\
    ~ & WNUT17 & 6 & 3,394 & 1,008 & 1,287 \\
    ~ & CrossNER\_AI & 13 & 100 & 350 & 431 \\
    ~ & CrossNER\_literature & 11 & 100 & 400 & 416 \\
    ~ & CrossNER\_music & 12 & 100 & 380 & 465 \\
    ~ & CrossNER\_politics & 8 & 199 & 540 & 650 \\
    ~ & CrossNER\_science & 16 & 200 & 450 & 543 \\
    \midrule
    \multirow{8}*{RE} & ACE05 & 6 & 10,051 & 2,420 & 2,050 \\
    ~ & ADE corpus & 1 & 3,417 & 427 & 428 \\
    ~ & CoNLL04 & 5 & 922 & 231 & 288 \\
    ~ & GIDS & 4 & 8,526 & 1,417 & 4,307 \\
    ~ & kbp37 & 18 & 15,917 & 1,724 & 3,405 \\
    ~ & NYT & 24 & 56,196 & 5,000 & 5,000 \\
    ~ & SciERC & 7 & 1,861 & 275 & 551 \\
    ~ & semeval RE & 10 & 6,507 & 1,493 & 2,717 \\
    \midrule
    \multirow{2}*{EE} & ACE05 & 33 & 19,216 & 901 & 676 \\
    ~ & CASIE & 5 & 11,189 & 1,778 & 3,208 \\
    \bottomrule
    \end{tabular}
    }
    \caption{\label{sft2-stat}
Statistics of the specific domain IE.}
\end{table}

\section{Details of Result Post-processing}
After the output codes are generated, we obtain the extraction results based on some regular expressions. To ensure the prediction results are more standardized and credible, two extra post-processing operations are added.

\paragraph*{Superclass Induction.} For the NER task, during the schema understanding phase, we have learned $29,177$ entity schemas, while the test dataset only contains $391$
schemas. For specific categories, our model may provide more detailed answers which are not in the dataset schema. For example, when it comes to the entity ``Harvard University'', our model tends to classify it as a \texttt{``\textcolor{darkgreen}{University}''}, while the ground truth labels it as an \texttt{``\textcolor{darkgreen}{Organization}''}. In such cases, we employ an upper-level recursive method to address this issue. Specifically, for the predicted entity, we perform Superclass Induction based on its position in the relationship tree in Wikidata. If the entity type of its upper-level concept
matches the entity type in the ground truth, we consider the entity prediction to be correct. 

\paragraph*{Type and Text Filtering.} For NER, RE, and EE tasks, if the model predicts a type that is not defined in the dataset schema and cannot be derived through superclass induction, or if an argument appears in the EAE task that is not present in the schema, we filter out such cases when calculating metrics. Additionally, if the model predicts text that does not appear in the sentence, we also filter it out.

\section{Implementation Details}
\label{implementation_details}
\paragraph*{Schema Understanding Phase.} The model is trained using AdamW~\cite{loshchilov2018adamw} optimizer with $\beta_1$ = $0.9$, $\beta_2$ = $0.95$, $\epsilon$ = $10^{-8}$ . We set the peak learning rate to $5 \times 10^{-6}$, and use a cosine learning rate schedule with warmup ratio of $0.1$, and decay final learning rate down to $10\%$ of the peak learning rate. To mitigate overfitting, we incorporated a weight decay of $0.1$ and a gradient clipping of $1.0$. We configure the context length to $2048$ and the global batch size to $1$M tokens, with the maximum training step capped at $4500$.

\paragraph*{Schema Following Phase.} We apply the LoRA~\cite{hu2021lora} method to all non-embedding linear layers for schema following. During this phase, we configure the LoRA rank and alpha parameters to $32$ and $64$, respectively, and set a dropout rate of $0.1$ to prevent overfitting. We still use the AdamW optimizer along with a cosine learning rate scheduler as in the schema understanding phase. The model undergoes $510$K training samples, with a learning rate of $3 \times 10^{-4}$, a global batch size of $256$, and a warmup ratio of $0.03$.

\paragraph*{Refinement Phase.} In the refinement phase, we employ a parameter configuration that is largely identical to the one used during the schema following phase. However, given the richer and more varied task-type data available during the refinement stage, we opt for a greater number of training iterations. Specifically, we conduct training over three epochs, cumulatively training on $1.9$M samples.


\newpage

\definecolor{darkorange}{RGB}{255, 140, 0}
\definecolor{lightgreen}{RGB}{145, 204, 117}
\definecolor{lightyellow}{RGB}{250, 200, 88}
\definecolor{lightred}{RGB}{238, 102, 102}
\definecolor{lightblue}{RGB}{115, 192, 222}

\definecolor{dkgreen}{rgb}{0,0.6,0}
\definecolor{gray}{rgb}{0.5,0.5,0.5}
\definecolor{mauve}{rgb}{0.58,0,0.82}

\newtcolorbox{promptbox}[2][Prompt]{
colback=black!5!white,
arc=5pt, 
boxrule=0.5pt,
fonttitle=\bfseries,
title=#1, 
before upper={\small}, fontupper=\fontfamily{ptm}\selectfont,
colframe=#2, 
}

\lstset{
    frame=single,
    language=Python,
    aboveskip=2mm,
    belowskip=2mm,
    xleftmargin=2mm,
    showstringspaces=false,
    keywordstyle=\color{blue},
    commentstyle=\color{dkgreen},
    stringstyle=\color{mauve},
    tabsize=3,
    numbers=left,
    numbersep=7pt,
    basicstyle={\small\ttfamily},
    backgroundcolor=\color[RGB]{245,245,244},
    numberstyle={\color[RGB]{0,192,192}\tiny},
    breaklines=true,
    breakatwhitespace=true,
    columns=flexible,
}

\onecolumn

\section{Cases of KnowCoder Training Data}

Here, we outline the cases that we have picked out from the KnowCoder-Dataset.

\subsection{Instance Code in Schema Understanding Phase}
\begin{promptbox}[\vspace{2pt}\large NER Task\vspace{2pt}]{entity}
\begin{lstlisting}
# Extract the entities from the following sentence.
sentence = "Lalita Yauhleuskaya competed at the 2008 Summer Olympics."

from Entities import Human

results = [
	Human("Lalita Yauhleuskaya")
]
\end{lstlisting}
\end{promptbox}

\bigskip
\bigskip
\bigskip

\begin{promptbox}[\vspace{2pt}\large RE Task\vspace{2pt}]{relation}
\begin{lstlisting}
# Extract the relations from the following sentence.
sentence = "Gzim Istrefi plays for Carlstad United BK."

from Entities import Human, AssociationFootballClub
from Relations import MemberOfSportsTeam

results = [
    MemberOfSportsTeam(
        Human("Gzim Istrefi"),
        AssociationFootballClub("Carlstad United BK")
    )
]
\end{lstlisting}

\end{promptbox}

\bigskip
\bigskip
\bigskip

\begin{promptbox}[\vspace{2pt}\large EE Task\vspace{2pt}]{event}
\begin{lstlisting}
# Extract the events from the following sentence.
sentence = "Jamsilsaenae station is adjacent to Sports Complex station which is on the Seoul Subway Line 2. The Sports Complex station is in the direction of Inner Ring Road and is located near Gangnam station."

from Entites import Entity
from Events import AdjacentStation

results = [
    AdjacentStation(
        connecting_line=[Entity("Seoul Subway Line 2")],
        towards=[Entity("Gangnam station")],
        direction=[Entity("Inner Ring Road")]
    )
]
\end{lstlisting}

\end{promptbox}

\newpage

\subsection{Instruction-tuning Code in Schema Following Phase}

\begin{promptbox}[\vspace{2pt}\large NER Task\vspace{2pt}]{entity}
\begin{lstlisting}
class Entity:
    """
    The base class for all entities.
    """
    def __init__(self, name: str):
        self.name = name

class Person(Entity):
    """
    Description: being that has certain capacities or attributes constituting personhood.
    Examples: patients, Donald Trump, children, women, user, patient, Trump, President Trump, Barack Obama, people
    """
    pass

class Nationality(SocialGroup):
    """
    Description: A legal identification of a person in international law, establishing the person as a subject, a national, of a sovereign state.
    Examples: American, British, Americans, German, French, English, Japanese, Russian, Australian, Indian
    """
    pass

class TvShow(Entity):
    """
    Description:
    Examples: Game of Thrones, The Walking Dead, American Idol, Modern Family, Saturday Night Live, Doctor Who, House, The Tonight Show, Mad Men, Arrested Development
    """
    pass

"""
This is an object-oriented programming task: some Entity Classes are defined above. Please instantiate all the corresponding Entity Objects in the following sentence.
"""
sentence = ``I enjoyed the series `Professional Master Chef' on television and I was struck by something the judges said when commenting about two of the semi-finalists. They had been highly impressed with the dishes the chefs had presented and Michel Roux Junior remarked that, despite their very obvious skill, neither chef exhibited any arrogance or conceit. Monica Galetti replied that they didn't need to, because their work spoke for them. ''
\end{lstlisting}
\begin{lstlisting}
results = [
    TvShow("Professional Master Chef"),
    Person("Michel Roux Junior"),
    Person("Monica Galetti")
]
\end{lstlisting}
\end{promptbox}

\newpage

\begin{promptbox}[\vspace{2pt}\large RE Task\vspace{2pt}]{relation}
\begin{lstlisting}
class Entity:
    """
    The base class for all entities.
    """
    def __init__(self, name: str):
        self.name = name

class Relation:
    """
    The base class for all relations.
    """
    def __init__(self, head_entity: Entity, tail_entity: Entity):
        self.head_entity = head_entity
        self.tail_entity = tail_entity

class PlaceOfBirth(Relation):
    """
    Description: Most specific known (e.g. city instead of country, or hospital instead of city) birth location of a person, animal or fictional character.
    Examples: (Australian, London), (Muhammad, Mecca), (Augustus, Rome), (Tiberius, Rome), (Mozart, Salzburg), (Charles II, London), (Sima Zhao, China), (Frederick the Great, Berlin), (Julius Caesar, Rome), (Queen Myeongui, Goryeo)
    """
    def __init__(self, head_entity: Entity, tail_entity: Entity):
        super().__init__(head_entity=head_entity, tail_entity=tail_entity)

class Population(Relation):
    """
    Description: Number of people inhabiting the place; number of people of subject.
    Examples: (civil parish, 201), (Sao Pedro, 201), (Machame Kusini, 13,572), (Sao Joao, 201), (unincorporated community, 15), (unincorporated community, 94), (unincorporated community, 25), (Mardekheh-ye Kuchek, 197), (Pain Halu Sara, 701), (Marenj, 1,055)
    """
    def __init__(self, head_entity: Entity, tail_entity: Entity):
        super().__init__(head_entity=head_entity, tail_entity=tail_entity)

class LocatedIn(Relation):
    """
    Description:
    Examples: (National Register of Historic Places, United States), (Ontario, Canada), (Sao Paulo, Brazil), (Victoria, Australia), (census-designated place, United States), (New South Wales, Australia), (California, United States), (Andes, Peru), (FAA, United States), (Norwegian, Norway)
    """
    def __init__(self, head_entity: Entity, tail_entity: Entity):
        super().__init__(head_entity=head_entity, tail_entity=tail_entity)

"""
This is an object-oriented programming task: some Relation Classes and related Entity Classes are defined above. Please instantiate all the corresponding Relation Objects in the following sentence.
"""
sentence = ``Kurush is a mountain village located in the Dokuzparinsky District, in southern Dagestan. Situated at 2480-2560 m above sea level depending on the source , it is the highest continuously inhabited settlement of the Greater Caucasus and of Europe as well as the southernmost settlement in Russia. As of 2015, Kurush had a population of 813.''
\end{lstlisting}
\begin{lstlisting}
results = [
    LocatedIn(Entity("Kurush"), Entity("Dokuzparinsky District")),
    LocatedIn(Entity("Dokuzparinsky District"), Entity("Dagestan")),
    Population(Entity("Kurush"), Entity("813"))
]
\end{lstlisting}
\end{promptbox}

\newpage

\begin{promptbox}[\vspace{2pt}\large ED Task\vspace{2pt}]{event}
\begin{lstlisting}
class Event:
    """
    The base class for all events.
    """
    def __init__(self, trigger: str, arg_names, *args):
        self.trigger = trigger
        self.arguments = {}
        for arg_name, arg_values in zip(arg_names, args):
            self.arguments[arg_name] = arg_values

class GroupMembership(Event):
    """
    Description: Organization, club or musical group to which the subject belongs.
    Examples: singer, music, musician, play, concert, performance, singing, sang, sung, sing,
    """
    def __init__(self, trigger: str, *args):
        arg_names = ["start", "role", "end", "group", "member"]
        super().__init__(trigger=trigger, arg_names=arg_names, *args)

class OlympicMedalHonor(Event):
    """
    Description: The honor associated with winning an Olympic medal.
    Examples: medal, gold, winner, win, silver, competition, bronze, victory, player, compete,
    """
    def __init__(self, trigger: str, *args):
        arg_names = ["event", "country", "medalist", "medal", "olympics"]
        super().__init__(trigger=trigger, arg_names=arg_names, *args)

class Education(Event):
    """
    Description: Educational institution attended by subject.
    Examples: school, professor, coach, graduate, student, study, master, education, pupil, lecturer,
    """
    def __init__(self, trigger: str, *args):
        arg_names = [
            "start_date",
            "degree",
            "end_date",
            "institution",
            "student",
            "specialization",
            "major_field_of_study",
        ]
        super().__init__(trigger=trigger, arg_names=arg_names, *args)

class Marriage(Event):
    """
    Description: The subject has the object as their spouse (husband, wife, partner, etc.).
    Examples: wife, married, husband, marriage, wedding, marry, couple, spouse, mistress, divorce,
    """
    def __init__(self, trigger: str, *args):
        arg_names = ["spouse", "location_of_ceremony", "type_of_union", "to", "from"]
        super().__init__(trigger=trigger, arg_names=arg_names, *args)

"""
This is an object-oriented programming task: some Event Classes are defined above. Please instantiate all the corresponding Event Objects in the following sentence.
"""
sentence = "Thomas Lincoln on June 12, 1806 married Nancy Hanks in the Richard Berry home."
\end{lstlisting}
\begin{lstlisting}
results = [
    Marriage("married")
]
\end{lstlisting}
\end{promptbox}

\newpage

\begin{promptbox}[\vspace{2pt}\large EAE Task\vspace{2pt}]{darkorange}
\begin{lstlisting}
class Entity:
    """
    The base class for all entities.
    """
    def __init__(self, name: str):
        self.name = name

class Event:
    """
    The base class for all events.
    """
    def __init__(self, trigger: str):
        self.trigger = trigger

class Education(Event):
    """
    Description: Educational institution attended by subject.
    """
    def __init__(
        self,
        trigger: str,  # Examples: school, professor, coach, graduate, student, study, master, education, pupil, lecturer,
        start_date: List[Entity],
        degree: List[Entity],
        end_date: List[Entity],
        institution: List[Entity],
        student: List[Entity],
        specialization: List[Entity],
        major_field_of_study: List[Entity],
    ):
        super().__init__(trigger=trigger)
        self.start_date = start_date
        self.degree = degree
        self.end_date = end_date
        self.institution = institution
        self.student = student
        self.specialization = specialization
        self.major_field_of_study = major_field_of_study

"""
This is an object-oriented programming task: some Event Classes are defined above. Please instantiate all the corresponding Event Objects in the following sentence. It is important to note that the triggers of the events are confirmed as follows: "graduate" is the trigger of event type "Education".
"""
sentence = "Albert J. Herberger (born c. 1933) is a Vice Admiral of the United States Navy, and the first United States Merchant Marine Academy graduate to attain the rank."
\end{lstlisting}
\begin{lstlisting}
results = [
    Education(
        trigger="graduate",
        institution=[Entity("United States Merchant Marine Academy")],
        student=[Entity("Albert J. Herberger")]
    )
]
\end{lstlisting}
\end{promptbox}

\end{document}